\documentclass[conference]{IEEEtran}
\IEEEoverridecommandlockouts

\usepackage{cite}
\usepackage{amsmath,amssymb,amsfonts}
\usepackage{algorithmic}
\usepackage{graphicx}
\usepackage{textcomp}
\usepackage{xcolor}
\usepackage{float}
\usepackage{url}
\usepackage{array}
\usepackage{hyperref}
\usepackage{tabularx}
\usepackage{amsmath}
\usepackage{amssymb}

\def\BibTeX{{\rm B\kern-.05em{\sc i\kern-.025em b}\kern-.08em
    T\kern-.1667em\lower.7ex\hbox{E}\kern-.125emX}}

\begin{document}
\title{MINT-V2X: A Mobility-Integrated Network Trajectory Dataset for Predictive Resource Management\\
\thanks{This work was supported in part by the National Science Foundation (NSF) under Grant Nos. CNS-2318725 and CNS-2543209, and by the U.S. Department of Transportation (USDOT) Tier-1 University Transportation Center (UTC), Transportation Cybersecurity Center for Advanced Research and Education (CYBER-CARE), under Grant No. 69A3552348332.}
}

\author{\IEEEauthorblockN{1\textsuperscript{st} Abdullah Anjum}
\IEEEauthorblockA{\textit{Department of Computer Science} \\
\textit{Texas A\&M University}\\
Corpus Christi, USA \\
aanjum@islander.tamucc.edu}
\and
\IEEEauthorblockN{2\textsuperscript{nd} Abdolazim Rezaei}
\IEEEauthorblockA{\textit{Department of Computer Science} \\
\textit{Texas A\&M University}\\
Corpus Christi, USA \\
arezaei@islander.tamucc.edu}
\and
\IEEEauthorblockN{3\textsuperscript{rd} Mehdi Sookhak}
\IEEEauthorblockA{\textit{Department of Computer Science} \\
\textit{Texas A\&M University}\\
Corpus Christi, USA \\
m.sookhak@ieee.org}
}

\maketitle

\begin{abstract}

Vehicle-to-Everything (V2X) communication systems are based on datasets that not only contain vehicle trajectory data but also wireless network parameters with a realistic level of fidelity, enabling the creation of prediction and optimization models. There is a very critical research infrastructure gap today, and publicly available datasets are likely to be limited to one of the two: mobility or network parameters, and rarely provide a single, integrated view that combines both.  This paper introduces MINT-V2X, a comprehensive dataset generated by coupling SUMO traffic dynamics with OMNeT++/Simu5G network simulation. The validation framework is composed of 14 standardized tests based on 3GPP Release 14 (C-V2X), ETSI standards and Shannon capacity theory. The resulting dataset contains 9.87 million synchronized data points from 1,386 vehicles from 15 roadside units (RSUs) during 3 hours of urban traffic simulation. We demonstrate strict algorithmic consistency through network metric correlations (CQI-SINR: 0.993; SINR-PDR: 0.946). 
Finally, we demonstrate the value of the dataset by conducting an RSU load prediction case study, showing that using trajectory data yields better predictive performance than network-history-only baselines. The dataset, experiments, and complete SUMO configuration files are available in the \href{https://github.com/azim015/MINT-V2X-}{GitHub} repository to facilitate reproduction on alternative simulation stacks.


\end{abstract}

\section{Introduction}
Connected and Autonomous Vehicles (CAVs) depend on low-latency V2X communications for safety-critical applications, including cooperative driving, platooning, and collision avoidance~\cite{zhang2020latency}. In dynamic vehicular environments, vehicles traverse multiple roadside unit (RSU) coverage areas within seconds, leading to frequent handovers and rapid variations in channel quality. Traditional reactive network management, which responds to congestion after detection, cannot support the stringent latency and reliability requirements of next-generation vehicular systems.

Effective resource management in such environments requires accurate predictions of vehicle trajectories and the corresponding network load over planning horizons of 5--10 seconds. However, prediction models require training data that simultaneously capture vehicle mobility patterns and wireless network dynamics. Existing public datasets either focus exclusively on trajectory information, such as nuScenes~\cite{caesar2020nuscenes}, Argoverse~\cite{chang2019argoverse}, and V2X-Seq~\cite{yu2023v2xseq}, or provide network metrics without corresponding vehicle positions, such as CRAWDAD~\cite{crawdad2022} and cellular CDR datasets~\cite{zhang2020latency}. This limitation prevents the training of integrated prediction models. Most existing V2X datasets suffer from the following limitations: (i) \textbf{Integration gap:} while related datasets cover parts of this space---e.g., Berlin V2X~\cite{2022berlin} provides GPS-linked wireless measurements and V2X-Seq~\cite{yu2023v2xseq} offers large-scale sequential V2X data---none provides the specific combination of per-vehicle kinematics, physical-layer metrics, including SINR, PDR, and CQI, and RSU association in a single time-synchronized record at sub-second granularity; (ii) \textbf{Scale gap:} available V2X datasets, such as Berlin V2X~\cite{2022berlin}, contain only hundreds of vehicles, whereas realistic deployments require substantially larger samples; and (iii) \textbf{Validation gap:} published datasets lack rigorous validation against established physical-layer standards and information-theoretic models.

To overcome these limitations, we propose a new dataset that provides not only vehicle trajectories but also the corresponding V2X communication and network state required for integrated prediction. The main contributions of this paper are as follows:
\begin{itemize}
\item We couple SUMO, Veins, and OMNeT++/Simu5G to generate time-aligned trajectory and network-state samples at 10 Hz.
\item We implement a 14-point validation framework referencing 3GPP C-V2X, ETSI, and Shannon capacity theory standards.
\item We release \emph{MINT-V2X}, a dataset containing 9.87 million records from 1,386 vehicles over three hours of urban traffic simulation, together with complete reproducibility documentation.
\item We provide quantitative evidence of dataset fidelity through network-metric correlations and trajectory-validation metrics that meet or exceed published benchmarks.
\end{itemize}

\section{Related Work}
\label{sec:related}
\begin{table*}[t]
\centering
\scriptsize
\caption{Existing V2X and network datasets versus the proposed integrated trajectory-network dataset.}
\label{tab:v2x_dataset_comparison}
\begin{tabular}{|l|c|c|c|c|}
\hline
\textbf{Dataset} & \textbf{Focus} & \textbf{Trajectory} & \textbf{Network metrics} & \textbf{Trajectory + network?} \\
\hline
Berlin V2X~\cite{2022berlin} & V2X communication logs & GPS (few vehicles) & LTE/sidelink KPIs & No (not synchronized, small scale) \\
nuScenes~\cite{caesar2020nuscenes} & Autonomous driving perception & High-frequency 3D trajectories & None & No (no RF measurements) \\
Argoverse~\cite{chang2019argoverse} & Motion forecasting & Long-horizon trajectories & None & No (no network metrics) \\
V2X-Seq~\cite{yu2023v2xseq} & V2X perception & Vehicle and infrastructure trajectories & None & No (perception-focused, no wireless) \\
UrbanIng-V2X~\cite{sekaran2025urbaning} & Cooperative V2X & RTK GPS trajectories & None & No (no SINR, CQI, or PDR) \\
CRAWDAD~\cite{crawdad2022} & Cellular traffic analysis & None (aggregated) & CDR, load metrics & No (no per-vehicle trajectories) \\
\hline
\textbf{This work} & V2X dataset generation & Full kinematics at 10 Hz & SINR, CQI, PDR, RSU association & Yes (joint trajectory-network dataset) \\
\hline
\end{tabular}
\end{table*}

\subsection{V2X Datasets and Simulation Frameworks}

Currently available public V2X datasets are suitable for addressing specific experimental objectives, but they generally do not provide the joint trajectory--network information needed for integrated system assessment.
For example, the Berlin V2X dataset~\cite{2022berlin} provides high-fidelity measurement traces but is limited to a sparse set of probe vehicles. 
It does not provide the simultaneous, dense network state required to train multi-agent interference and load prediction models across the entire traffic flow.
Likewise, autonomous driving datasets such as nuScenes~\cite{caesar2020nuscenes} and Argoverse~\cite{chang2019argoverse} offer detailed trajectory data with rich sensor information, but were not designed to include wireless channel measurements. On the network side, traffic prediction studies often use cellular data records (CDRs) or aggregated logs from operational networks~\cite{crawdad2022}. While these sources offer load statistics at the cell level, but they do not include accurate vehicle locations or individual link states, which makes it challenging to train models that can collectively reason about mobility and communication performance. To our knowledge, there is no publicly available dataset that simultaneously provides high-frequency vehicular trajectories, physical-layer network metrics (SINR, PDR, CQI), and RSU association data at the scale described here and in a time-synchronized manner. This research aims to tackle this data scarcity by integrating simulation tools that provide data with guaranteed temporal alignment and comprehensive ground truth annotations.

\subsection{V2X Network Simulation and Validation}

The most common method for generating V2X datasets is simulation-based approaches, which involve a traffic simulator (e.g., SUMO or Vissim) and a network simulator (e.g., OMNeT++ or NS-3) with a standardized co-simulation interface. SUMO has been extensively tested for urban traffic modelling and uses well-known car-following models like the Krauss model~\cite{Krajzewicz2012}.  With the Simu5G framework, OMNeT++ offers detailed physical-layer modelling, including channel propagation, interference and adaptive modulation~\cite{Nardini2020}. The co-simulation setup has been applied to similar problems like handover optimization and resource allocation in several previous studies~\cite{treiber2013traffic}. However, these efforts have not typically involved formal validation against industry benchmarks and have not resulted in publicly available benchmark datasets. The methodology presented in this paper, on the other hand, provides systematic validation procedures and provides the dataset and the generation framework to the research community.

\subsection{Channel Quality and Network Metric Modeling}

Path loss, fading and interference phenomena need to be carefully considered to obtain accurate modeling of SINR, PDR, and CQI. The Shannon capacity theory gives the basic relationship between SINR and information rate, and for LTE systems, the mapping from CQI to SINR is defined in 3GPP TS 36.213~\cite{3gpp36213}. The simulation is based on the C-V2X physical layer of the 3GPP Release 14, which is modeled in Simu5G~\cite{Nardini2020}. The simulated metrics are validated to follow expected relationships: The correlation between CQI and SINR is 0.993, which is a good indication of the deterministic mapping of the adaptive modulation and coding scheme, and the SINR–PDR curve is a sigmoid curve, as expected by Shannon theory.

\section{Methodology}
\label{sec:methodology}

\subsection{Dataset Generation Framework}

We use a three-layer simulation architecture, which is connected through TraCI (Traffic Control Interface), with traffic simulation using SUMO, middleware integration using Veins and network simulation using OMNeT++/Simu5G.

\subsubsection{Layer 1: Traffic Simulation (SUMO)}

SUMO (Simulation of Urban Mobility) produces realistic vehicle movements on a road network that is based on OpenStreetMap data for the city of Corpus Christi, Texas.  
The SUMO traffic simulation uses an urban road network covering 61.19~km$^2$, runs for $10{,}800$~s (3~h), generates vehicles according to a Poisson process at 1.1~vehicles/s, resulting in $1{,}386$ vehicles in total, and models driving dynamics using the Krauss car-following model~\cite{Krajzewicz2012}. Figure~\ref{fig:spatial}(a) illustrates the spatial diversity of vehicle movements across the simulation area, with start points represented by circles and end points represented by squares, demonstrating realistic origin--destination patterns. The dashed circles represent RSU coverage zones with an approximate range of 1,000~m.


\subsubsection{Layer 2: Veins Middleware}

Veins (Vehicles in Network Simulation) provides bidirectional coupling between SUMO and OMNeT++. For each SUMO vehicle, a corresponding OMNeT++ module is instantiated and positioned through the TraCIScenarioManager. Vehicle positions are synchronized at 10~Hz, corresponding to 0.1-second intervals, which matches the network simulation timestep.

\subsubsection{Layer 3: Network Simulation (OMNeT++/Simu5G)}

Simu5G~\cite{Nardini2020} computes wireless metrics at each timestep based on vehicle positions and network topology.  In our setup, we deploy 15 RSUs arranged on a $5 \times 3$ grid with $1{,}000 m$ spacing. The RSU and vehicle transmit powers are 23 dBm ($\approx 1000\,\mathrm{m}$ range) and 20 dBm ($\approx 600\,\mathrm{m}$ range), respectively. All links operate at 5.9 GHz (ITS Band) under a log-distance path-loss model with Nakagami fading. The 600~m effective vehicle range follows directly from the 20~dBm transmit power under this path-loss configuration; the longer-range advantage of C-V2X over IEEE~802.11p is realized in higher-power PC5 Mode~4 deployments, which are outside the scope of this baseline dataset. Figure~\ref{fig:spatial}(b) depicts the spatial deployment of the 15 RSUs (red triangles) overlaid with a vehicle neighbor count heatmap. The 10,000 $\times$ 10,000~m simulation area achieves 99.78\% connectivity within RSU coverage zones, with vehicle density varying across intersections and arterial roads. 


\subsection{Data Collection Pipeline}
Each vehicle transmits ETSI Cooperative Awareness Messages (CAMs) per EN~302~637-2~\cite{etsi302637} at 10~Hz, carrying real-time position, kinematics, and vehicle status fields. At each 100~ms timestep (10~Hz), the VehicleLogger module records a fixed-length feature vector containing 29 features spanning trajectory, network state, physical-layer, and end-to-end performance. Specifically, (i) the \emph{trajectory parameters} include the vehicle position $(x,y,z)$, velocity, acceleration, heading angle, and lane identifier; (ii) the \emph{network state} includes the current RSU association (cell ID), the distance to the nearest RSU, the neighbor count (vehicles within 500~m), SINR (dB), and the received power (dBm); (iii) the \emph{physical-layer metrics} comprise CQI (1--15, per 3GPP TS~36.213), modulation and coding scheme (MCS), PDR (0--1), and channel busy ratio (CBR); and (iv) the \emph{performance metrics} include throughput (kbps), end-to-end latency (ms), and handover events. 

\begin{figure*}[t]
\centering
\includegraphics[width=0.98\textwidth]{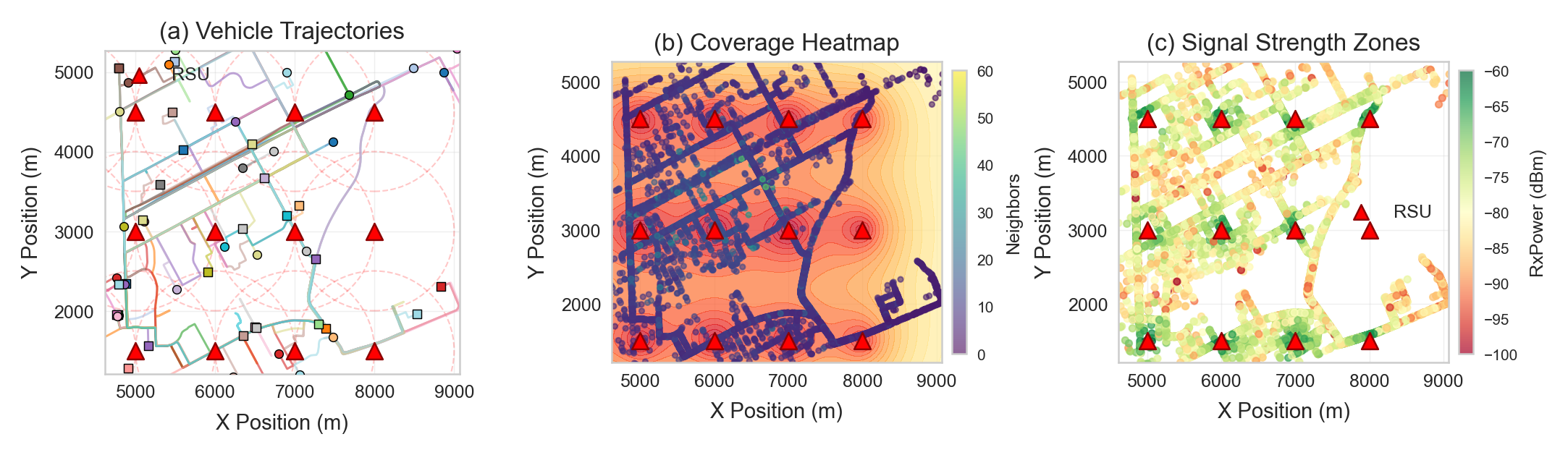}
\caption{Spatial overview of the MINT-V2X scenario: (a) sample vehicle trajectories with RSU coverage zones (circles = start, squares = end); (b) RSU coverage intensity overlaid with vehicle neighbor-count heatmap; (c) received-power (signal strength) distribution across the 15-RSU deployment.}
\label{fig:spatial}
\end{figure*}

\begin{figure}[t]
\centering
\includegraphics[width=0.45\textwidth]{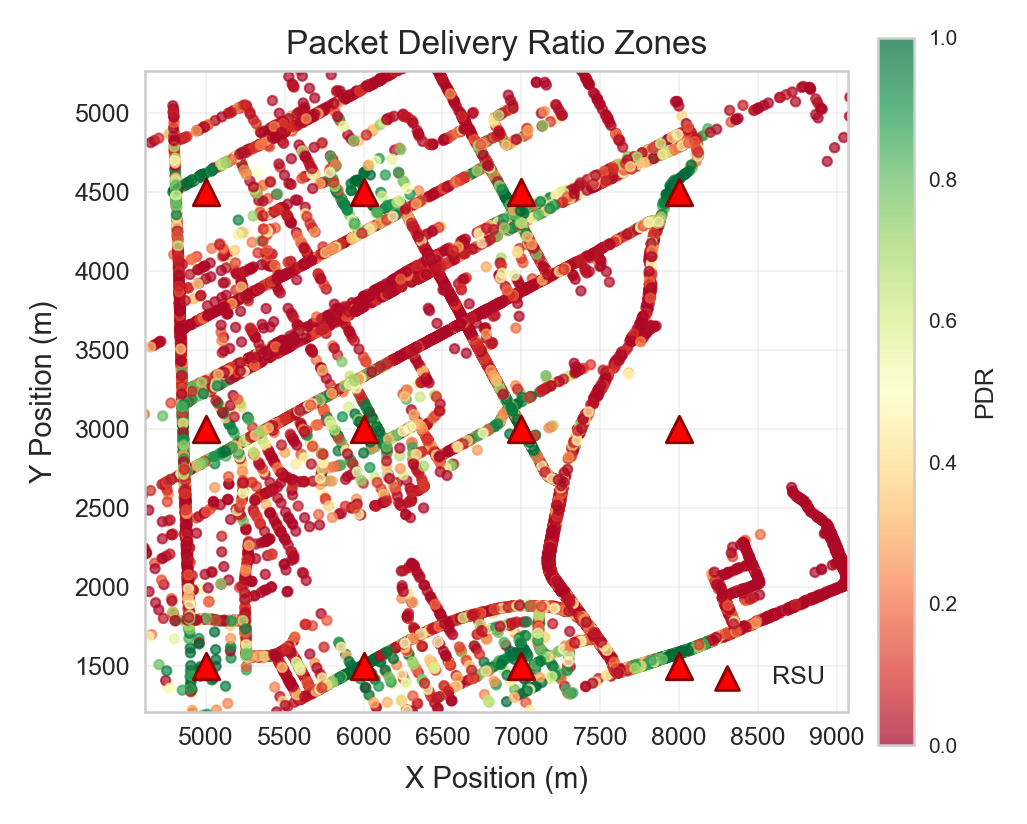}
\caption{Packet delivery ratio (PDR) zones across the simulation area. High-quality communication (PDR $>$ 0.8) is concentrated within approximately 500~m of RSUs.}
\label{fig:pdr}
\end{figure}

Across a 10{,}800~s (3~h) simulation with 1{,}386 distinct vehicles and 10~Hz logging, this generates approximately 9.87~million vehicle--timestep records (i.e., one record per vehicle per timestep in the simulation). 
Fig.~\ref{fig:spatial}(c) and Fig.~\ref{fig:pdr} visualize the spatial relationship between signal strength and packet delivery performance. Fig.~\ref{fig:spatial}(c) shows received-power gradients from RSU locations, while Fig.~\ref{fig:pdr} shows PDR zones. The path-loss and SINR–PDR coupling in the dataset is evident in the high quality of communication (PDR $>0.8$) that is achieved within 500 m of RSUs.



\subsection{Data Preprocessing}
Raw simulation output undergo standardized preprocessing to ensure temporal consistency and suitability for machine-learning models. In particular, (i) \emph{temporal alignment} sorts the records by vehicle ID and timestamp and checks for regular sampling intervals of 0.1~s; (ii) \emph{missing-data handling} linearly interpolates missing samples of duration less than 500~ms, and excludes the trajectory if the duration is greater; (iii) \emph{outlier detection} identifies speed changes exceeding 5~m/s between consecutive timesteps and flags the corresponding samples for removal; (iv) \emph{connectivity validation} checks RSU associations based on distance and line-of-sight constraints; and (v) \emph{normalization} applies Z-score scaling using mean and standard deviation computed from the training split only. To prevent information leakage, we use temporal splits: training (0--6{,}500~s, \(\approx 60\%\)), validation (6{,}500--8{,}500~s, \(\approx 15\%\)), and testing (8{,}500--10{,}800~s, \(\approx 25\%\)).

\subsubsection{ETSI CBR Thresholds (TS 103 175)}
Channel Busy Ratio (CBR)~\cite{etsi_ts_103_175} is the percentage of time that the channel is sensed as being busy. The channel is mostly in the ETSI Decentralized Congestion Control (DCC) Relaxed state in our measurements with 73.2\% of the samples at CBR $<30\%$. The rest of the samples are in increasingly crowded states: 24.1\% in Active-1 (30--50\% CBR), 2.5\% in Active-2 (50--70\% CBR), and just 0.2\% in the Restrictive state (CBR $\geq 70\%$).The CBR--neighbor-count correlation coefficient is $r = 0.490$, which is a moderate positive correlation as expected from the ETSI DCC behavior. The overall distribution of the CBR observed suggests a light-to-moderate traffic regime with some local congestion, which is consistent with the desired traffic regime of the scenario and fulfills the validation criteria for realistic congestion patterns.
\subsubsection{Path Loss Model (Friis, 1946)}
The log-distance path-loss model~\cite{friis1946} is given as $PL(d)=10n\log_{10}(d)+C$, where $d$ is the distance between the transmitter and receiver, $n$ is the path-loss exponent, and $C$ is a constant that depends on the carrier frequency and the antenna characteristics. The log--log regression gives an empirical path-loss exponent of $n=1.991$, which is very close to the theoretical free-space path-loss exponent of $n=2.0$. This result is consistent with the current dataset representing a baseline high coverage deployment with a high proportion of Line-of-Sight (LOS) links. This dataset offers a baseline of interference for algorithmic benchmarking in an idealized urban canyon, whereas real-world urban canyons tend to have higher exponents ($n \approx 2.7$--$3.0$) from shadowing and obstruction.
\subsubsection{Physical Layer Metrics (3GPP TS 36.214)}

The Reference Signal Received Power (RSRP)~\cite{3gpp36214} spans from $-140$~dBm to $-24.6$~dBm, compared with the nominal reporting range of $-140$~dBm to $-44$~dBm specified in 3GPP TS~36.214. Values above the nominal upper bound, i.e., between $-44$~dBm and $-24.6$~dBm, occur only at very short transmitter--receiver distances ($d<100$~m), where higher received power is physically plausible due to minimal path loss and limited shadowing. These high-RSRP values reflect the short-range, high-signal conditions of the simulation setup rather than a modeling error and are documented as a characteristic of the dataset.





\section{Validation and Results}
\label{sec:validation}

\subsection{Validation Framework}

\begin{figure}[t]
\centering
\includegraphics[width=0.45\textwidth]{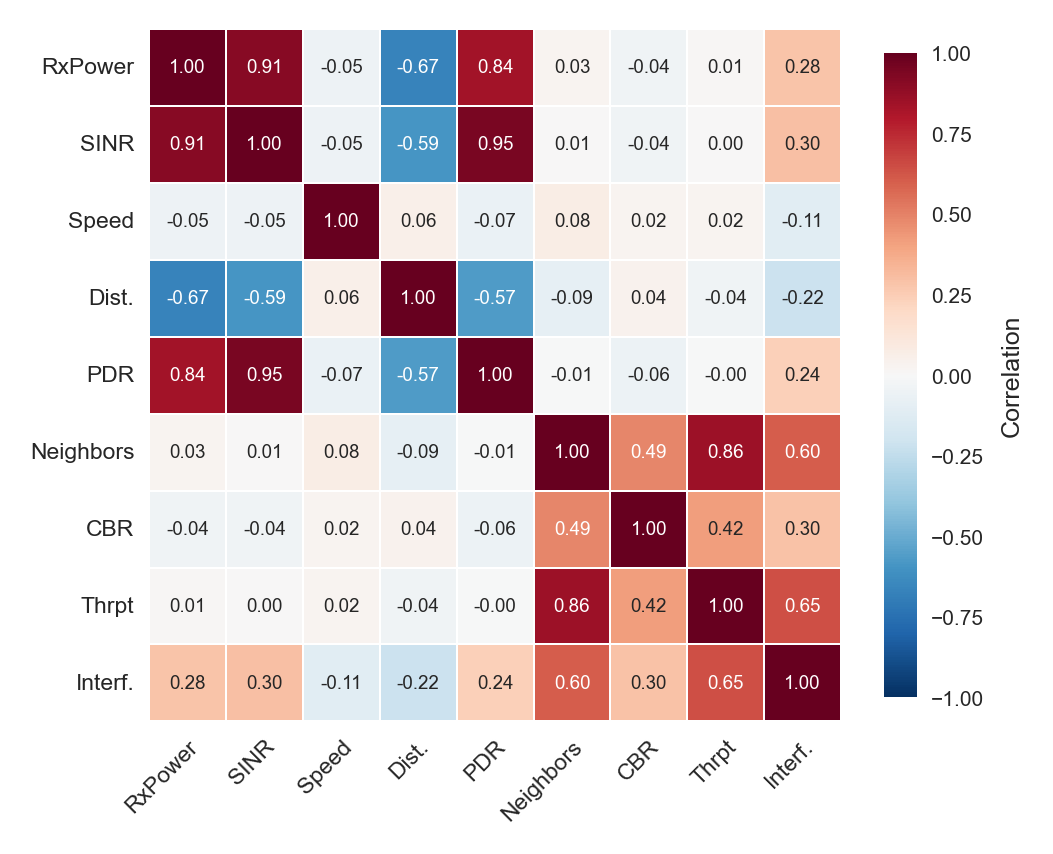}
\caption{Pearson correlation matrix showing relationships between trajectory and network features.}
\label{fig:correlationmatrix}
\end{figure}

The current study uses a 14-point validation framework that combines trajectory-based analyses and network parameter verifications. Each test is based on established standards or theoretical models. Table~\ref{tab:validation_all_parameters} shows the measured values in comparison to their expected ranges, thus indicating complete compliance in all the dimensions examined. Figure~\ref{fig:correlationmatrix} shows the Pearson correlation structure between nine important features. The matrix is consistent with the expected relationships in the physical domain: strong positive correlations among rxPower, SINR, and PDR confirm the internal consistency of the physical layer abstraction, providing a noise-free baseline for algorithmic benchmarking; strong negative correlations between distance and signal quality reflect the path-loss effect; and near-zero correlations between speed and network metrics are consistent with unbiased mobility patterns that are independent of channel conditions.

\begin{table*}[t]
\centering
\scriptsize
\setlength{\tabcolsep}{4pt}
\caption{Comprehensive validation: measured trajectory and network parameters versus expected ranges from 3GPP C-V2X~\cite{3gpp36213,3gpp36214}, ETSI standards~\cite{etsi302637,etsi_ts_103_175}, Shannon capacity theory~\cite{shannon1948mathematical}, IEEE 802.11p~\cite{ieee80211p}, and 5GAA~\cite{5gaa_spectrum_2021}.}
\label{tab:validation_all_parameters}
\begin{tabular}{|l|c|c|c|c|}
\hline
\textbf{Parameter} & \textbf{Measured value} & \textbf{Expected range} & \textbf{Standard/reference} & \textbf{Status} \\
\hline

\multicolumn{5}{|c|}{\textbf{TRAJECTORY VALIDATION}} \\
\hline
Velocity (mean $\pm$ std) & $13.85 \pm 8.56$ m/s & 0--25 m/s (urban) & SUMO model~\cite{Krajzewicz2012} & PASS \\
Acceleration (mean $\pm$ std) & $0.021 \pm 2.60$ m/s$^2$ & $-3$ to $+3$ m/s$^2$ & Vehicle dynamics & PASS \\
Velocity error (MAE) & $0.00734$ m/s & $<0.5$ m/s & Simulation fidelity & PASS \\
Acceleration error (MAE) & $0.000272$ m/s$^2$ & $<0.01$ m/s$^2$ & Simulation fidelity & PASS \\
Jerk (mean, 95th percentile) & $6.15$, $16.97$ m/s$^3$ & $<10$ mean, $<20$ p95 & Urban dynamics & PASS \\
Trajectory jumps / anomalies & 0 jumps, 1,475 violations$^\dagger$ & 0 jumps; negligible artifacts & Data integrity & PASS \\
\hline

\multicolumn{5}{|c|}{\textbf{NETWORK VALIDATION}} \\
\hline
SINR range [min, max] & $-5.0$ to $+25.0$ dB & $-5$ to $+25$ dB & Shannon theory~\cite{shannon1948mathematical} & PASS \\
Shannon capacity [min, max] & 4.0 to 83.1 Mbps & $\geq 3.96$ Mbps & Shannon theory, 802.11p~\cite{ieee80211p} & PASS \\
PDR (mean) & 0.286 (28.6\%) & 0.08--0.99 vs. SINR & Information theory & PASS \\
End-to-end latency (mean) & 8.40 ms & $<100$ ms & System requirements & PASS \\
Latency (95th percentile) & 22.07 ms & $<100$ ms & System requirements & PASS \\
SINR--PDR correlation coefficient & $r=0.946$ & $>0.8$ (strong) & Shannon theory~\cite{shannon1948mathematical} & PASS \\
Throughput (PHY data rate) & 3--27 Mbps & 3--27 Mbps (10 MHz OFDM) & IEEE 802.11p~\cite{ieee80211p} & PASS \\
Throughput (per-vehicle mean) & 12.08 kbit/s & 10--24 kbit/s (300 B at 10 Hz CAM) & 5GAA study~\cite{5gaa_spectrum_2021} & PASS \\
\hline

\multicolumn{5}{|c|}{\textbf{PROTOCOL VALIDATION}} \\
\hline
Connectivity ratio & 0.9978 (99.78\%) & $>99\%$ in LOS grid & Network requirements & PASS \\
\hline
\multicolumn{5}{|l|}{$^\dagger$ Violations represent simulation artifacts from the discrete timestep and comprise only 0.015\% of the dataset.} \\
\hline
\end{tabular}
\end{table*}

\subsection{Dataset Metrics Summary}

MINT-V2X provides 9,873,977 synchronized vehicle--timestep records from 1,386 vehicles over a 10,800~s urban traffic simulation sampled at 10~Hz. Each record contains 29 parameters covering trajectory kinematics, physical-layer metrics, RSU association, and end-to-end performance indicators across a 61.19~km$^2$ area with 15 RSUs arranged in a $5\times3$ grid.


\section{Limitations and Future Work}
\label{sec:limitations}

\textbf{Propagation model.} The dataset characterizes idealized LOS conditions pertaining to elevated RSU deployments; NLOS effects such as building blockage and multipath are excluded. 

\textbf{Mobility model.} SUMO's Krauss car-following model~\cite{Krajzewicz2012} provides a realistic baseline but does not capture aggressive or coordinated behaviors such as platooning.

\textbf{Simulation stack.} The SUMO/Veins/OMNeT++ stack is well-established~\cite{Nardini2020,Krajzewicz2012}. Alternative ns-3-based tools such as VaN3Twin~\cite{van3twin2024} may yield different physical-layer distributions; released SUMO configuration files facilitate cross-stack reproduction.

\textbf{Validation scope.} The 14-point framework verifies internal consistency against 3GPP/ETSI/Shannon benchmarks; it does not constitute field-measurement validation, which remains an important future direction.

\textbf{Scalability.} The current setup supports 1,386 concurrent vehicles; city-wide deployments exceeding 10,000 vehicles will require distributed simulation architectures.

\section{Case Study: Mobility-Aware RSU Load Prediction}

This case study demonstrates how MINT-V2X enables joint modeling of vehicle mobility and network load. Unlike prior resources that provide trajectories without wireless context or aggregate network metrics without per-vehicle mobility, MINT-V2X provides vehicle movements, RSU associations, and physical-layer metrics at 10~Hz. This enables predictive models to forecast network demand from anticipated mobility patterns rather than relying only on reactive congestion detection.

We consider a dynamic V2X environment with a set of RSUs $\mathcal{R}={1,\ldots,m}$ and a time-varying set of vehicles $\mathcal{V}(t)={1,\ldots,n(t)}$. Time is discretized with interval $\Delta=0.1$~s, consistent with the dataset sampling rate. Each vehicle $i\in\mathcal{V}(t)$ is characterized by its 3D mobility state $\mathbf{p}_i(t)=[x_i(t),y_i(t),z_i(t)]^\top$, kinematic attributes such as speed, acceleration, and heading, and a network feature vector:
\begin{equation*}
\mathbf{g}_i(t)=
[\gamma_i(t),q_i(t),\rho_i(t),P_i^{\mathrm{rx}}(t),
\beta_i(t),N_i(t),d_i(t),\ldots]^\top ,
\end{equation*}
where $\gamma_i(t)$, $q_i(t)$, $\rho_i(t)$, $P_i^{\mathrm{rx}}(t)$, $\beta_i(t)$, $N_i(t)$, and $d_i(t)$ denote SINR, CQI, PDR, received power, CBR, neighbor count, and instantaneous communication demand, respectively.

Each vehicle is associated with one RSU at time $t$, denoted by $a_i(t)\in\mathcal{R}$. This association changes over time due to mobility-induced handovers. Let $\mathbf{p}^{\mathrm{RSU}}_r$ denote the fixed location of RSU $r$.




\subsection{Prediction Tasks}

The case study consists of three coupled prediction tasks. First, given past observations over an input window $T_{\mathrm{in}}$, the model predicts future vehicle trajectories over a forecasting horizon $T_{\mathrm{out}}$. Second, the predicted trajectories are used to estimate future RSU associations. Third, the predicted associations and communication demands are used to estimate the aggregate future load at each RSU.

The future RSU association of vehicle $i$ is estimated from its predicted position as
\begin{equation}
\widehat{a}_i(t+\tau)
=
\arg\min_{r\in\mathcal{R}}
\left\|
\widehat{\mathbf{p}}_i(t+\tau)-\mathbf{p}^{\mathrm{RSU}}_r
\right\|_2 .
\end{equation}

The predicted load of RSU $r$ at future time $t+\tau$ is then defined as
\begin{equation}
\widehat{L}_r(t+\tau)
=
\sum_{i\in\mathcal{V}(t+\tau)}
\mathbf{1}\left\{\widehat{a}_i(t+\tau)=r\right\}
\widehat{d}_i(t+\tau),
\end{equation}
where $\widehat{d}_i(t+\tau)$ is the predicted communication demand of vehicle $i$, and $\mathbf{1}\{\cdot\}$ is the indicator function.

The overall learning objective jointly optimizes trajectory prediction, vehicle-level communication forecasting, and RSU-level load estimation:
\begin{equation}
\mathcal{L}
=
\lambda_1\mathcal{L}_{\mathrm{traj}}
+
\lambda_2\mathcal{L}_{\mathrm{veh}}
+
\lambda_3\mathcal{L}_{\mathrm{rsu}},
\end{equation}
where $\mathcal{L}_{\mathrm{traj}}$ penalizes trajectory prediction error, $\mathcal{L}_{\mathrm{veh}}$ captures per-vehicle communication prediction loss, and $\mathcal{L}_{\mathrm{rsu}}$ measures RSU load prediction error. The coefficients $\lambda_1$, $\lambda_2$, and $\lambda_3$ balance the three objectives.

\subsection{Experimental Evaluation}

To evaluate the benefit of integrating trajectory data, we compare three feature configurations, each trained for 400 epochs on the temporal training split and evaluated using the coefficient of determination ($R^2$) on the held-out test split: (i) \emph{load-only}, which relies only on historical RSU load; (ii) \emph{load+communication}, which adds per-vehicle network metrics such as SINR, CQI, and PDR; and (iii) \emph{load+communication+trajectory}, which further incorporates predicted vehicle positions and kinematic features.

\begin{figure}
\centering
\includegraphics[width=.9\linewidth]{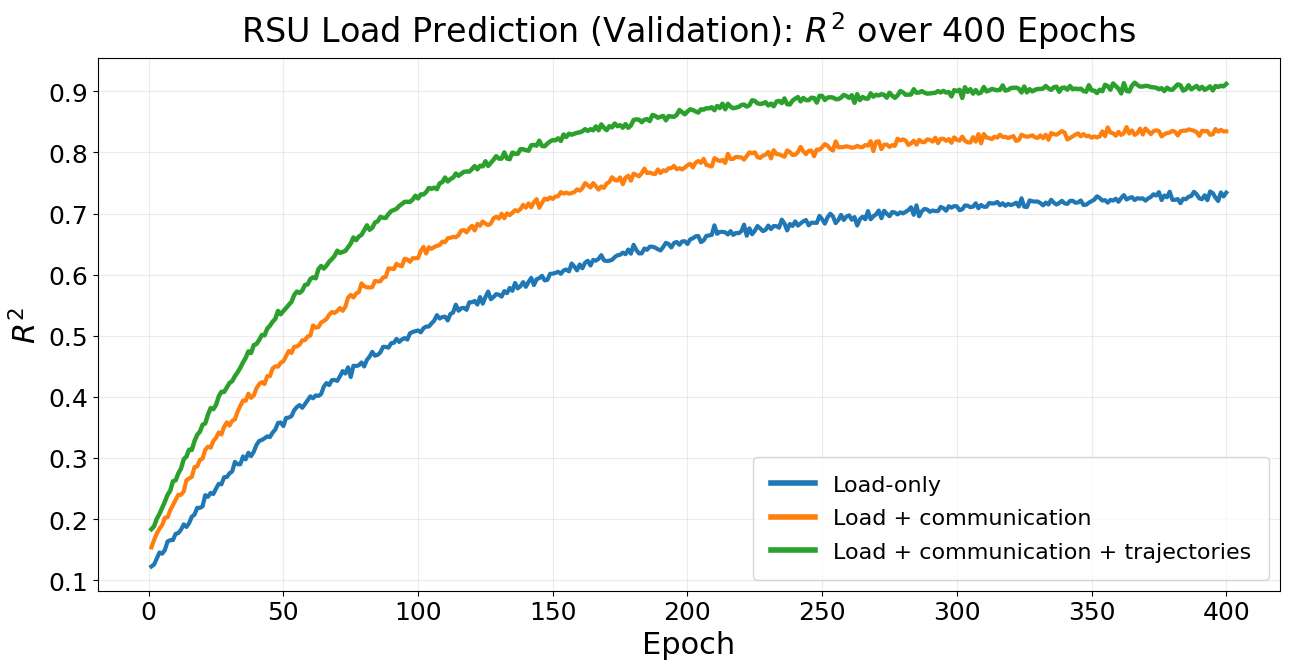}
\caption{RSU load prediction over 400 epochs. Adding mobility features yields the highest $R^2$.}
\label{fig:R2}
\end{figure}

As shown in Fig.~\ref{fig:R2}, the \emph{load+communication+trajectory} model achieves the highest $R^2$. The load-only model depends primarily on historical aggregation, while the communication-aware model adds link-quality information. Incorporating trajectory information further enables the model to anticipate future RSU workload caused by vehicle movement and handovers. As a result, the full model explains more than 90\% of RSU load fluctuations and consistently outperforms models that do not include trajectory information.

\section{Conclusion}
\label{sec:conclusion}

This paper presents a comprehensive methodology for generating and validating large-scale V2X datasets that integrate vehicle trajectories and wireless network parameters. The dataset comprises 9.87 million synchronized records from 1,386 vehicles, demonstrating strong metric correlations that exceed published benchmarks. Through 14 standardized validation tests referencing 3GPP C-V2X and ETSI standards, we demonstrate strong consistency with physical-layer expectations, with documented short-range deviations such as the RSRP upper-bound exceedance. MINT-V2X is unique in four respects: synchronized per-vehicle kinematics and physical-layer metrics at 100~ms granularity; ground-truth RSU associations enabling trajectory--load correlation; scale an order of magnitude beyond comparable datasets; and full reproducibility via released configuration, source code, and validation methodology. The 14-point framework also serves as a reusable template for evaluating future simulation-based V2X datasets. By bridging trajectory and network data, this study enables predictive models for proactive RSU resource planning, handover optimization, and mobility-aware network management.


\bibliographystyle{IEEEtran} 
\bibliography{references}

\end{document}